# Hand Pose Classification Based on Neural Networks

*Rashmi Bakshi*

*TU Dublin*

**Abstract**

In this work, deep learning models are applied to a segment of a robust hand-washing dataset that has been created with the help of 30 volunteers. This work demonstrates the classification of presence of one hand, two hands and no hand in the scene based on transfer learning. The pre trained model; simplest NN from Keras library is utilised to train the network with 704 images of hand gestures and the predictions are carried out for the input image. Due to the controlled and restricted dataset, 100% accuracy is achieved during the training with correct predictions for the input image. Complete hand washing dataset with dense models such as AlexNet for video classification for hand hygiene stages will be used in the future work.

## 1  Introduction

Deep learning is a branch of machine learning which is based on an artificial neural network (ANN) having an ability to mimic the behaviour of a human brain. Artificial neurons are the fundamental component for building ANNs. ANN consists of multiple hidden layers with multiple hidden units (neurons) or nodes [1]. It is an emerging approach and has been widely applied in traditional artificial intelligence domains such as semantic parsing, transfer learning, computer vision, natural language processing and more [2]. Over the years, Deep learning has gained increasing attention due to the significant low cost of computing hardware and access to high processing power (eg-GPU units) [2]. Conventional machine learning techniques were limited in their ability to process data in its natural form. For decades, constructing a machine learning system required domain expertise and fine engineering skills to design a feature extractor that can transform the raw data (example: pixel values of an image) into a feature vector, which is passed to a classifier for pattern recognition [3]. Deep learning models learn features directly from the data without the need for building a feature extractor. "Deep" usually refers to the number of hidden layers in the neural network. Traditional neural networks contain 2-3 hidden layers where as deep networks can have as many as 150 hidden layers and so it requires high computing power [4]. In this work, simple NN model is implemented from Keras library on a hand-washing dataset that involves different hand-washing gestures.

## 2  Successful applications based on deep learning solutions

As deep learning continues to mature, there are many deep learning applications ranging from language recognition, self-driving cars to medical research classification of CT images for lung nodules [5]. Few applications in the field of Computer vision and pattern recognition that gained popularity with the use of deep learning models are:

- Automatically colorize gray-scale images: Convolution neural network is utilized to jointly extract local and global features from an image and then fuse them together to perform the final colorization of black and white images [6].

- Generating automatic natural language description of images and their regions: Deep neural network model is developed that associates segments of text with the region of the image that they describe in order to create a label for the images [7].
- Text recognition: Recognizing and retrieving text in natural scene images- text based image retrieval [8].
- Visually indicated sounds: Recurrent neural network is developed to predict sound features from silent videos by analysing the visual scene [9].
- Deep photo style transfer: Photographic style transfer seeks to transfer the style of a reference style photo onto another image. The deep network is built to transfer colours from the input image and avoids spatial distortions [10].

# 3   Convolution Neural Network (CNN) Architecture

CNN is a type of deep learning model for processing data that has a grid pattern such as images and primarily used in the field of pattern recognition within images. CNN consists of three types of layers. They are Convolution layers, pooling layers and fully connected layers. Convolution and pooling layers perform feature extraction while a fully connected layer maps the extracted features into the final output [5, 11]. A CNN architecture is formed when these layers are stacked together.

**Convolution layers:** It is responsible to extract low-level features such as edges, colour from the input image with the use of a convolution filter, which traverses through the entire image. With added layers, the architecture adapts to high-level features building a network that understands the overall images in the data set.

**Pooling layers:** It is responsible for reducing the spatial size of the given input thereby reducing the number of subsequent learning parameters; flatten the image and transform into a 1D array of numbers (or vectors).

**Fully connected layers:** It is responsible for connecting one or more fully connected layers, also known as dense layers, in which every input is connected to every output by a learnable weight [5].

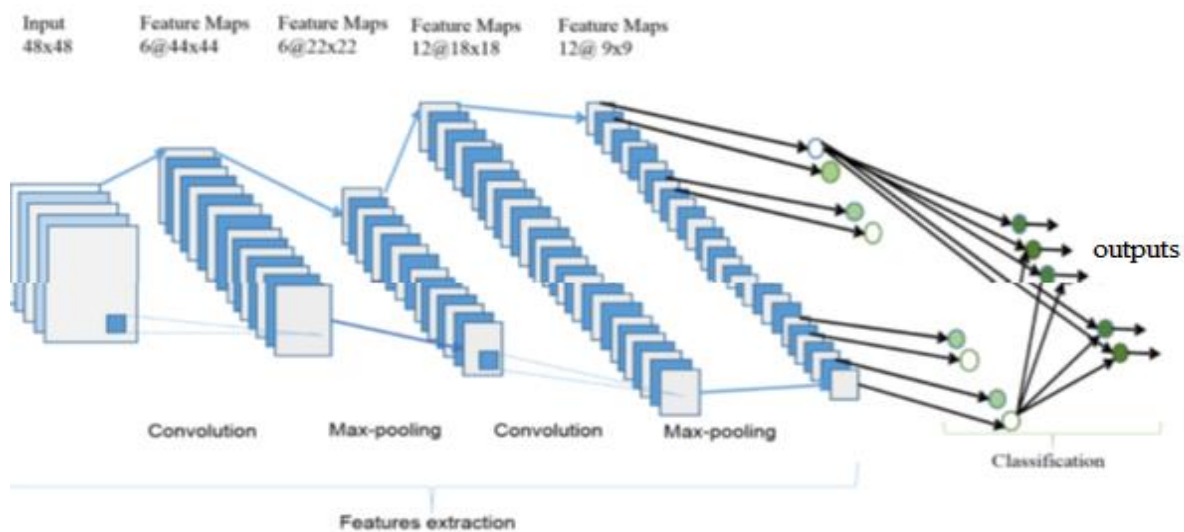

Figure 3.1: The overall architecture of CNN [1]

# 4   Transfer learning

The main challenge of this work is the acquisition of "large-scale data sets". Deep learning models essentially require thousands of data samples and heavy computational resources such as GPU for accurate classification and prediction analysis. However, there is a branch of machine learning, popularly known as "transfer learning" that does not necessarily requires large amounts of data for evaluation. Transfer learning is a machine learning technique wherein a model, which is developed for one task, is reused for the second related task. It refers to the situation where "finding" of one setting is exploited to improve the optimisation in another setting [12]. Transfer learning is usually applied to the new data set, which is generally smaller than the original data set used to train the pre-trained model. To give an example:

Hussain et al. applied transfer learning to train Caltech face data set with 450 face images with pre-trained model on ImageNet data set. Increasing the number of training steps (epochs) increased the classification accuracy but increased the training time as well. Computational power and time were the main limitations within the study [12].

Transfer Learning is an important tool in machine learning that can solve the basic problem of an insufficient training data. It tries to transfer the knowledge from the source domain to the target domain where training and test data is not required to be independent and identically distributed [13].

Yu-Chuan Su et al. applied transfer learning from image to video where the image corpus is weak labelled available in various social media such as Flicker, Instagram, and videos were manually labelled YouTube videos [14].

In this work, pre-trained models available from Keras library are utilised to classify the presence of hands in the scene. 704 images extracted from hand hygiene video recordings were segmented into three classes. This work is a preliminary analysis of applying transfer-learning paradigm to an existing data set. The future work will focus upon the video classification of various hand hygiene movements with increased amount of data with the help of pre-trained models available in Keras library.

# 5   Hand Gestures Dataset

A video-based dataset that consists of Hand-washing gestures were created with the help of 30 participants. The participants were instructed about the hand hygiene guidelines before the beginning of the data collection process. The participant's consent was obtained and only the hand movements were captured in order to maintain the anonymity. The video length for the hand washing activity was recorded for 25-30 seconds. Every hand-washing step was followed by a pause where in the participant was instructed to move their hands away from the camera. Video format for this data set is MP4 file with a size of range 40-60 MB and a frame rate of 29.84 frames/s. All of the six hand washing movements were recorded in one video for each participant. In addition to the hand-washing dataset, one-hand gestures such as linear hand movement and circular hand movement were captured in a separate video recording for all the participants. In this paper, preliminary classification is carried out to detect the presence of one hand, two hands, no hand (only background) from the dataset created.

# 6   Methodology

The video recordings were decomposed into images and three classes were created. The images in the classes were evenly distributed in order to avoid the bias during the training of the model. A small part of the dataset was applied in this work to test the accuracy of the model. It is very controlled and limited dataset in order to learn the techniques of transfer learning at the beginner's level. In future, robust dataset with large amount of data will be explored. Figure 6.1 elaborates the steps carried out for the training of the simplest NN model. Two python scripts were created, one for training the model with training set and validation set. Another script was written for making predictions on the new data. The scripts are adaptation from [15].

Class 1: No hand- 232 images, 76.6MB

Class 2: One hand- 233 images, 73.8MB

Class 3: Two hands- 239 images, 83.2MB

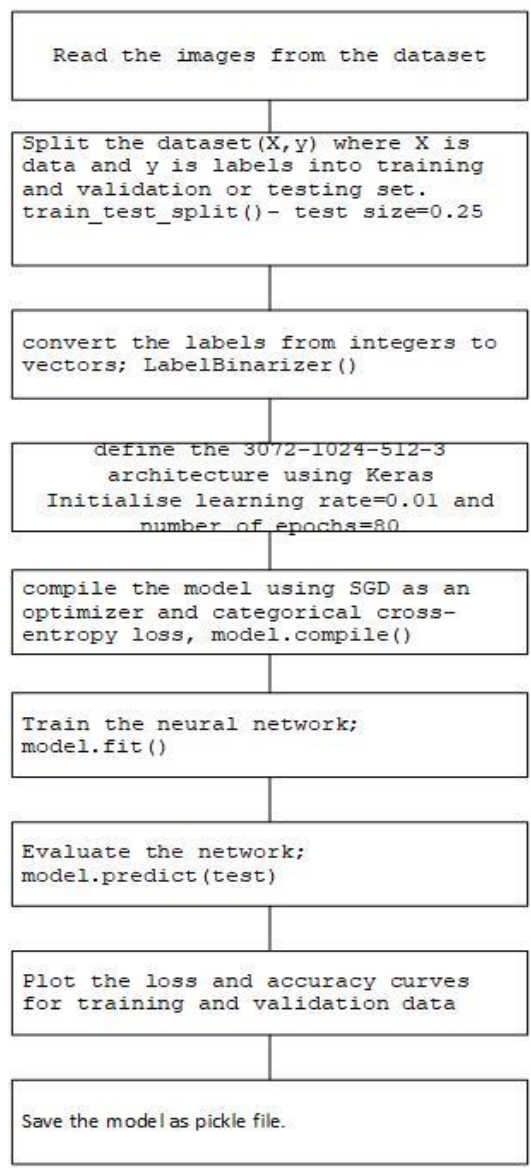

Figure 6.1: Basic workflow for training the neural network. [15]

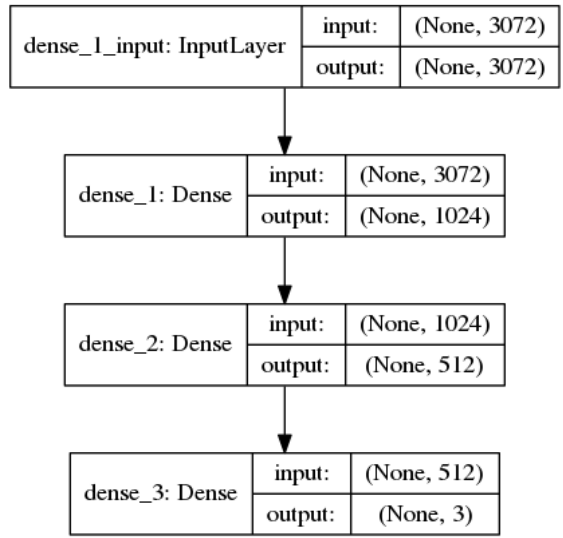

Figure 6.2: Keras Model architecture used in this work [15]

*Predictions.py*

```
Read the image; resize

Load the saved model ; pickle.loads()

Make a prediction on the image; model.predict(image)

Find the class label index with the largest corresponding probability;

Draw the class label + probability on the output image

Show the output image
```

# 7 Results

Figure 7.1 is the loss-accuracy plot for the training data and the validation data. As number of training steps (epochs) increase, the loss curve continues to fall for both sets. An accuracy score has reached to 1.0 by the last training step. 100% accuracy is achieved due to the lack of robust dataset and less amount of data used in this experiment; however transfer learning gives promising results and make correct predictions for the input images (Figure 7.2-7.4). In future, the complete hand washing dataset will be explored in addition to the use more advanced deep learning models such as VGGNet and AlexNet.

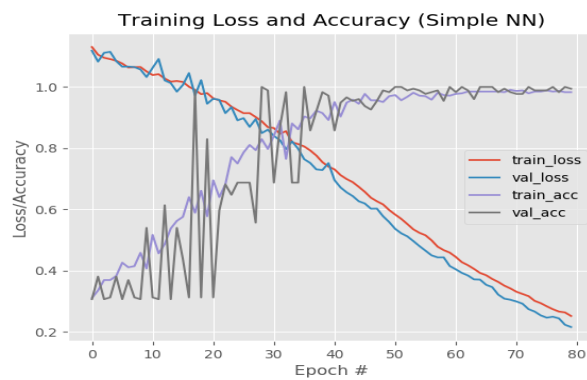

Figure 7.1: Loss-Accuracy curve for this experiment

Class predictions

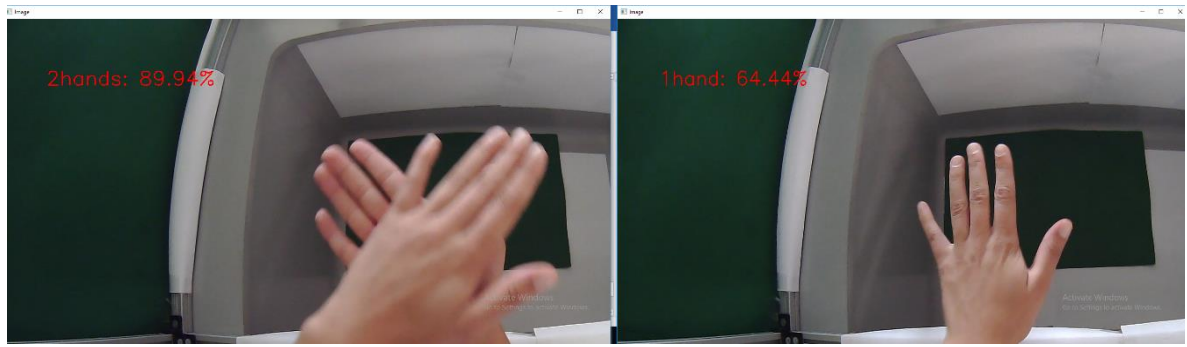

Figure 7.2: 2 hands predicted with 89.94% accuracy (L) and 1 hand predicted with 64.44% accuracy(R)

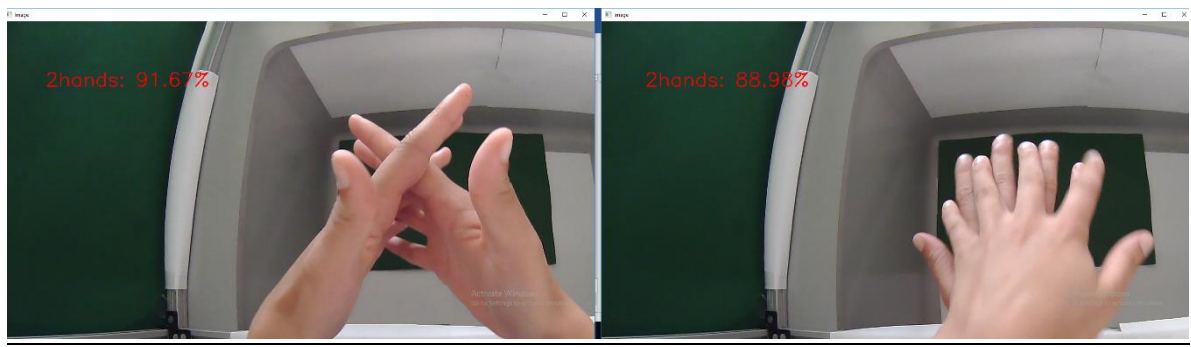

Figure 7.3: 2 hands predicted with 91.67% accuracy (L) and 2 hands predicted with 88.98% accuracy(R)

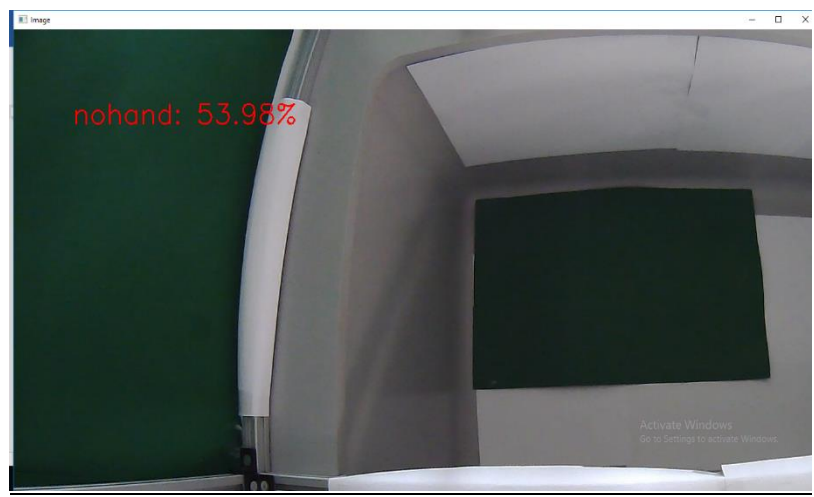

Figure 7.4: No hands predicted with 53.98% accuracy